\pdfoutput=1

\documentclass[11pt]{article}

\usepackage[final]{acl}

\usepackage{times}
\usepackage{latexsym}

\usepackage[T1]{fontenc}

\usepackage[utf8]{inputenc}

\usepackage{microtype}

\usepackage{inconsolata}

\usepackage{graphicx}

\usepackage[edges]{forest}
\usepackage{ccg}
\usepackage{amstext}
\usepackage{subcaption}  
\usepackage{booktabs}
\usepackage{algpseudocode}
\usepackage{adjustbox}
\usepackage{multicol}
\usepackage{algorithm}

\usepackage{booktabs}
\usepackage{tabularx}
\usepackage{xcolor}
\usepackage{caption}
\usepackage{amsmath}
\usepackage{multirow}
\usepackage{cleveref}
\usepackage{bm}
\Crefname{section}{\S}{\S\S}

\newcommand{\hidden}[1]{}
\def\1{\bm{1}}

\usepackage{array}
\definecolor{highlight}{RGB}{0, 102, 204}
\definecolor{derivation}{gray}{0.3}
\definecolor{highlightbg}{RGB}{255, 255, 153} 
\definecolor{typegray}{gray}{0.3}

\title{Which Word Orders Facilitate Length Generalization in LMs? \\ An Investigation with GCG-Based Artificial Languages}

\author{Nadine El-Naggar\thanks{Equal contribution.}\\
  \And
  Tatsuki Kuribayashi\footnotemark[1]\\
  Mohamed bin Zayed University of Artificial Intelligence \\
    \texttt{\{nadine.naggar, tatsuki.kuribayashi, ted.briscoe\}@mbzuai.ac.ae} \\
  \And
  Ted Briscoe \\}

\begin{document}
\maketitle
\begin{abstract}
 Whether language models (LMs) have inductive biases that favor typologically frequent grammatical properties over rare, implausible ones has been investigated, typically using artificial languages (ALs)~\cite{DBLP:conf/acl/WhiteC20,DBLP:conf/acl/KuribayashiUYOB24}.
In this paper, we extend these works from two perspectives.
First, we extend their context-free AL formalization by adopting Generalized Categorial Grammar (GCG) \cite{wood2014categorial}, which allows ALs to cover attested but previously overlooked constructions, such as unbounded dependency and mildly context-sensitive structures.
Second, our evaluation focuses more on the generalization ability of LMs to process unseen longer test sentences.
Thus, our ALs better capture features of natural languages and our experimental paradigm leads to clearer conclusions --- typologically plausible word orders tend to be easier for LMs to productively generalize.

\end{abstract}

\section{Introduction}
Attested natural languages (NLs) possess different grammatical properties, such as different word orders.
This naturally raises a question about what kind of language is easier for language models (LMs) to learn~\cite{cotterell+al.naacl18a,DBLP:conf/acl/MielkeCGRE19,DBLP:conf/acl/WhiteC20,borenstein-etal-2024-languages,arnett-bergen-2025-language}.
This question has even been extended to counterfactual, impossible languages~\cite{mitchell-bowers-2020-priorless,DBLP:conf/acl/KalliniPFMP24,DBLP:conf/acl/KuribayashiUYOB24}. 
Two related additional questions are why are some combinations of features typologically common and others rare~\cite{wals}, and what role if any can LMs play in exploring such questions~\cite{chomsky2023noam}.

To answer these questions, we need to understand how we can adequately measure the inductive bias of LMs over specific grammatical properties?
There are at least two challenges from both data and evaluation metric perspectives.
On the data side, NLs differ across a variety of dimensions, and thus isolating a specific grammatical factor for evaluation is challenging with NL data \cite{DBLP:conf/acl/MielkeCGRE19}. 
The use of artificial languages (ALs), instead, is a promising direction to enable more controlled experimental setups~\cite{DBLP:conf/acl/WhiteC20}, but ALs are often highly simplified and lack critical properties underlying NLs, such as context-free Dyck languages.
On the evaluation metric side, LM performance is often measured with perplexity (PPL) on the held-out dataset sampled from the same distribution (domain) as the training data.\footnote{We use the terms ``in-domain'' and ``out-of-domain'' just based on the length of the dataset in this study, while these are often relevant to more semantic differences of the data.}
An additional important aspect to be evaluated in language learning is, however, the ability to productively generalize to longer sentences from shorter stimuli, generally motivated by the argument of ``infinite use of finite means''.

In this paper, we advance this line of research on both data and evaluation sides.
For the data, we introduce an extensible approach to defining ALs, based on Generalized Categorial Grammars (GCGs)~\cite{wood2014categorial}.
Our framework can support the inclusion of mildly context-sensitive (indexed language) constructions, such as cross-serial dependencies, and a general approach to unbounded filler-gap dependencies, while maintaining diverse naturalistic constructions.
We exemplify this by extending the set of ALs in \citet{DBLP:conf/acl/WhiteC20} to include object relative clauses as one exemplar of unbounded dependencies.

For the evaluation metric, we target the generalization of LMs from shorter exposures during training to a longer test set.
That is, we train LMs on a set of shorter AL sentences and then evaluate their performance on the unseen, longer AL sentences.
We further introduce several evaluation perspectives on the generalization test set, including PPLs on specific challenging constructions that require proper syntactic generalization, such as unbounded dependencies, as well as grammatical judgment accuracy rather than holistic PPL scores.

In our experiments, following \citet{DBLP:conf/acl/WhiteC20} and \citet{DBLP:conf/acl/KuribayashiUYOB24}, we repeatedly evaluate LM's generalization ability, using different ALs with different word order configurations.
Here, we try to answer the question of which word order configurations facilitate LMs to better perform generalization to longer sentences and accurately make grammaticality judgments.
In particular, we are interested in whether typologically plausible word orders make it easier for LMs to perform productive linguistic generalization.

Our experimental results offer several novel findings.
First, out-of-domain evaluation with longer sentences makes LMs' inductive bias clearer, compared to preference across different word orders on in-domain  (same length as training data) evaluation. 
Second, stronger correlations between LM performance and typological distributions emerge once the scope is extended from in-domain to generalization based evaluation.
That is, typologically plausible word order tends to be easier for LMs to perform  \textit{generalization} to longer sentences, rather than just fitting to in-domain data.
Third, RNN's performance is overall better aligned with typological plausibility throughout our three experiments, than other architectures, such as Transformers, which supports that working memory constraints shape typologically frequent word orders in natural language~\cite{Hawkins1994-sw,futrell2020dependency,Hahn2020-sb}.

\section{Background}

\subsection{Artificial Language Learning}

ALs are often used in targeted evaluation of LMs. 
One line of research uses ALs to assess whether LMs can learn patterns corresponding to different levels of the Chomsky Hierarchy. \citet{DBLP:conf/coling/SomeyaYO24}, for instance, test whether LMs can learn regular, context-free, and context-sensitive languages, specifically those involving nested, long-distance, and cross-serial dependencies. 
Additional studies use context-free and mildly context-sensitive languages, like Dyck languages and $a^n b^n c^n$, to test how well LMs generalize to longer sequences \cite{suzgun2019lstm,weiss-et-al-2018-practical,el-naggar2022exploring}, and explore how different LM architectures correspond to various levels of the Chomsky Hierarchy \cite{deletang2022neural}. However, a key limitation is that many of these ALs are far removed from natural language, involving highly simplified vocabularies, unrealistic degrees of (self-)embedding and 
limited constructional variety.

Another area of research builds on the claim by \citet{chomsky2023noam} that neural LMs can learn both 
possible and impossible
human languages, making them unable to distinguish between the two. 
\citet{DBLP:conf/acl/KalliniPFMP24} constructed typologically impossible ALs by systematically permuting and modifying an English dataset, following \citet{DBLP:conf/naacl/RavfogelGL19}. Their experiments show that GPT-2 models struggle to learn these impossible ALs, 
which is inconsistent with the claims by \citet{chomsky2023noam}.
Still, it is difficult to pinpoint exactly which linguistic features make language learning more challenging, due to the complex, multidimensional nature of the modified input.

Inspired by \citet{ravfogel2018can},
\citet{DBLP:conf/acl/WhiteC20} use ALs generated by a probabilistic context-free grammar (PCFG) to study the inductive biases of LMs towards particular word orders. 
By defining six structural parameters that reverse the order of constituents in different syntactic rules, they generate a range of word order configurations and evaluate LSTM and Transformer performance across them. 
\citet{DBLP:conf/acl/KuribayashiUYOB24} extend this work by evaluating cognitively inspired LMs on the same ALs. 
However, due to the constraints of the PCFGs used, their ALs do not include several attested grammars or constructions, such as
Verb-Subject-Object (VSO) word order, and mildly context-sensitive constructions.
They also do not test LMs' generalization to longer sentences.

Concurrent works have also explored other formulations of ALs; for example, \citet{xu2025can} explored dependency-based corpus modification, \citet{hunter2025kallini} proposed ALs with constituency-based non-adjacency, and \citet{yang-etal-2025-anything} used multiple languages as a seed NLs to develop ALs.
Our previous work introduced GCG-based ALs, but did not test length generalization and focused on the replication and extension of existing PPL-based studies~\cite{el-naggar-etal-2025gcg}.

\subsection{Generalization to Longer Sentences}
Human language possesses the property of productive and systematic compositionality, where new sentences are formed from known basic components \cite{doi:10.1111/j.1755-2567.1970.tb00434.x,Chomsky1957-CHOSS-2}. 
Humans are able to produce and comprehend an open-ended number of sentences from early limited exposure to short ones. 
This indicates that humans are able to generalize during learning from short (in-domain) sentences to longer (out-of-domain) sentences.
There is a long-standing debate on whether neural network (NN) models can generalize productively and systematically \cite{fodor1988connectionism,baroni2020linguistic}.
Notably, the generalization of LMs to complex (potentially longer) sequences has been evaluated in a wide range of tasks, e.g., deductive reasoning~\cite{clark2021transformers,saparov2023testing}, arithmetic reasoning~\cite{kudo2023deep}, or programming~\cite{dziri2023faith}.

ALs are often used to evaluate LM's fundamental linguistic competence and their generalization ability to longer sentences, typically with, e.g.,  Dyck languages and $a^n b^n$($c^n$)~\cite{weiss-et-al-2018-practical,suzgun2019lstm,el-naggar2022exploring,el2023theoretical}. 
\citet{weiss-et-al-2018-practical}, for example, empirically test LSTM, GRU \cite{DBLP:conf/emnlp/ChoMGBBSB14}, and Elman RNN \cite{elman1990finding} LMs, and LSTMs learn $a^n b^n$ most effectively, but they eventually fail on longer sequences.
Similarly, \citet{suzgun2019lstm} empirically assess the ability of their LSTM models to learn Dyck languages effectively and generalize to longer sequences. 
However, they do not address whether this behavior is precise enough to generalize to sequences that are significantly longer.
\citet{el-naggar2022exploring} use Dyck languages to evaluate long-term generalization of counting on LSTM, ReLU and GRU models.
They use training and test sets of the same size and sequence length as \citet{suzgun2019lstm}, but additionally test their models on significantly longer sequences, and find that their models do not generalize effectively to these very long sequences.

Another commonly used AL for model generalization is SCAN \cite{lake2017generalization}.
They evaluate the models' generalizability to new combinations from familiar components, e.g., from ``jump'' and ``twice'' to  ``jump twice.'' 
Still, the mentioned ALs for generalization tests, including $a^n b^n$($c^n$) languages, Dyck languages, and SCAN, do not reflect many of the properties of attested NLs, and may not be adequate to evaluate inductive biases in realistic language learning scenarios.

\begin{figure*}[t]
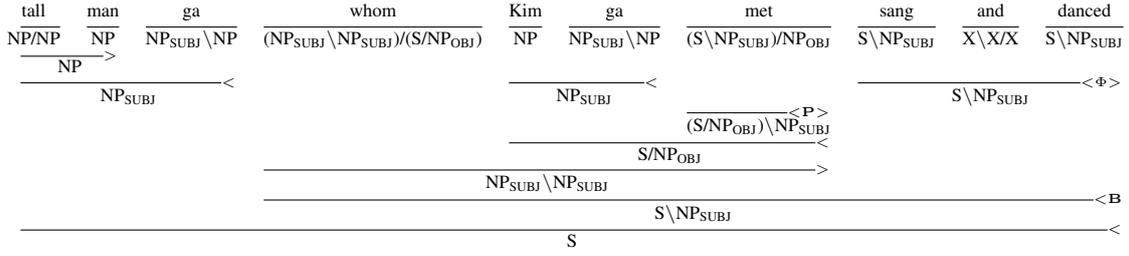

    \scriptsize{
    \begin{center}

        \deriv{10}{
    \text{tall}&\text{man}&\text{ga}&\text{whom}&\text{Kim}&\text{ga}&\text{met}&\text{sang}&\text{and}&\text{danced}\\
    \uline{1}&\uline{1}&\uline{1}&\uline{1}&\uline{1}&\uline{1}&\uline{1}&\uline{1}&\uline{1}&\uline{1}\\
    \text{NP/NP} & \text{NP}&\text{NP$_\text{SUBJ}$\bs NP}&\text{(NP$_\text{SUBJ}$\bs NP$_\text{SUBJ}$)/(S/NP$_\text{OBJ}$)}& \text{NP}&\text{NP$_\text{SUBJ}$\bs NP}&\text{(S\bs NP$_\text{SUBJ}$)/NP$_\text{OBJ}$}&\text{S\bs NP$_\text{SUBJ}$}&\text{X\bs X/X} & \text{S\bs NP$_\text{SUBJ}$}\\
    \fapply{2} & & & & & & & &\\
    \mc{2}{\text{NP}}& & & & & & & &\\
    \bapply{3}  & & \bapply{2} & & \conj{3}\\
    \mc{3}{\text{NP$_\text{SUBJ}$}}& &\mc{2}{\text{NP$_\text{SUBJ}$}}& & \mc{3}{\text{S\bs NP$_\text{SUBJ}$}}\\
    & & & & & & \permute{1} & & &\\
    & & & & & & \mc{1}{\text{(S/NP$_\text{OBJ}$)\bs NP$_\text{SUBJ}$}} & & &\\
     & & & &  \bapply{3} & & &\\
     & & & &  \mc{3}{\text{S/NP$_\text{OBJ}$}} & & &\\
    & & &   \fapply{4} & & &\\
    & & &   \mc{4}{\text{NP$_\text{SUBJ}$\bs NP$_\text{SUBJ}$}} & & &\\
    & & & \bcomp{7}\\
    & & & \mc{7}{\text{S\bs NP$_\text{SUBJ}$}}\\
    \bapply{10}\\
    \mc{10}{\text{S}}\\
    
    }
    \end{center}}
    \caption{Example of a sentence and its derivation.
    }
    \label{fig:derivation_sampling}
\end{figure*}

\subsection{Categorial Grammar}

A categorial grammar (CG) consists of a lexicon that assigns each word a basic or functor category, along with a set of rules that define how functor categories combine with basic categories in both syntax and semantics.
Slash notation is used to indicate the direction of the argument relative to the resulting category: for example, 
$\alpha / \beta$
denotes a functor that expects a 
$\beta$
to its right to form an expression of category 
$\alpha$.
Classical CG includes just two combinatory rules: \textbf{forward functional application} (a) and \textbf{backward functional application} (b), as shown below.
\begin{enumerate}

    \item[(a)] $\alpha/\beta \text{ } \beta \Rightarrow \alpha$
    \item[(b)] $\beta \text{ } \alpha \bs \beta \Rightarrow \alpha$
\end{enumerate}
Below is an example of forward and backward application using the English transitive verb "chased", which is the functor category $(S\bs NP)/NP$.
\begin{center}
\deriv{3}{

\text{Tom} & \text{chased} & \text{Jerry} \\
\uline{1} & \uline{1} & \uline{1}\\
\text{NP} & \text{(S\bs NP)/NP} & \text{NP}\\
 & \fapply{2} \\
 & \mc{2}{\text{S\bs NP}}\\
 \bapply{2} & \\
 \mc{2}{\text{S}} & \\
}
\end{center}

We use English examples to demonstrate rules and derivations.
In CG, the majority, if not all, of the variation across languages can be attributed to differences in the lexical categories assigned to words.

CG, which has the expressive power of context-free grammar (CFG), has been extended to combinatory categorial grammar (CCG) \cite{steedman1996surface}, and generalized categorial grammars (GCG) \cite{wood2014categorial} by introducing additional operations to combine categories.
One such operation is \textbf{composition}, which, like functional application, has forward (a) and backward (b) variants, shown below. 
\begin{enumerate}
    \item[(a)] $\alpha/\beta \text{ } \beta/\gamma \Rightarrow \alpha/\gamma$
    \item[(b)] $\beta\bs \gamma \text{ } \alpha\bs \beta \Rightarrow \alpha\bs \gamma$
\end{enumerate}
Composition (\textbf{B}) is demonstrated
below.
\begin{center}
    \deriv{4}{
    \text{child} & \text{in} & \text{school} & \text{studied}\\
    \uline{1} & \uline{1} & \uline{1} & \uline{1} \\
    \text{NP} & \text{(NP\bs NP)/NP} & \text{NP} & \text{S\bs NP}\\
     & \fapply{2} & \\
     & \mc{2}{\text{NP\bs NP}} & \\
     & \bcomp{3}\\
     & \mc{3}{\text{S\bs NP}}\\
     \bapply{4}\\
     \mc{4}{\text{S}}\\
    }
\end{center}
Another operation in GCG is \textbf{coordination} ($\Phi$), where 2 constituents with the same categories can be combined into a single one of the same type if they are separated by a conjunction.
This is demonstrated in the example below.
\begin{center}
\deriv{5}{
\text{Tom} & \text{and} &  \text{Jerry} & \text{caused} & \text{trouble}\\
\uline{1} & \uline{1} & \uline{1} & \uline{1 }& \uline{1}\\
\text{NP} & \text{CONJ} & \text{NP} &  \text{(S\bs NP)/NP} & \text{NP}\\
\conj{3} & & \\
\mc{3}{\text{NP}}&  \\
 & & &  \fapply{2} \\
 & & & \mc{2}{\text{S\bs NP}}\\
 & \bapply{4} \\
 & \mc{4}{\text{S}} \\
 }
\end{center}

We do not use CCG-style type raising, and instead use \textbf{permutation} from GCG due to its greater computational tractability.
We use cyclic permutation as defined by \citet{briscoe1997co,briscoe2000grammatical}, where the arguments to functor categories can be cyclically permuted while maintaining their directionality.
Formally:
$$(\alpha|\beta_{1})...|\beta_{n} \Rightarrow (\alpha|\beta_{n})|\beta_{1}$$
Permutation (P) is shown 
below:
\begin{center}
    \deriv{4}{
    \text{man} & \text{whom} & \text{I} & \text{met}\\
    \uline{1} & \uline{1} &\uline{1} &\uline{1} \\
    \text{NP} & \text{(NP\bs NP)/(S/NP)} & \text{NP} & \text{(S\bs NP)/NP} \\
     & & & \permute{1} \\
     & & & \mc{1}{\text{(S/NP)\bs NP}} \\
     & & \bapply{2} \\
     & & \mc{2}{\text{S/NP}} \\
     & \bapply{3}\\
     
     & \mc{3}{\text{NP\bs NP}}\\
     \bapply{4} \\
     \mc{4}{\text{NP}} \\
    }
\end{center}

We design our ALs based on the application, composition, coordination, and permutation rules defined above. 
Figure~\ref{fig:derivation_sampling} shows an example of the parse of an English-like AL sentence.

\begin{table*}[t]
\centering
\small
\tabcolsep 0.1cm
\begin{tabular}{llp{5.5cm}p{5.5cm}}
\toprule
Param. & Description                                     & 0 (head-final)                                                                                                                                                                                                & 1 (head-initial)                                                                                                                                                    \\
\cmidrule(r){1-1} \cmidrule(lr){2-2} \cmidrule(lr){3-3} \cmidrule(lr){4-4}
\texttt{S}      & Order of subject and verb                           & \begin{tabular}[c]{@{}l@{}}VI $\rightarrow$ S\textbackslash{}NP$_\text{SUBJ}$\\ VT $\rightarrow$ (S\textbackslash{}NP$_\text{SUBJ}$)$\mid$NP$_\text{OBJ}$\\ VCOMP $\rightarrow$ (S\textbackslash{}NP$_\text{SUBJ}$)$\mid$SCOMP\end{tabular} & \begin{tabular}[c]{@{}l@{}}VI $\rightarrow$ S/NP$_\text{SUBJ}$\\ VT $\rightarrow$ (S/NP$_\text{SUBJ}$)$\mid$NP$_\text{OBJ}$\\ VCOMP $\rightarrow$ (S/NP$_\text{SUBJ}$)$\mid$SCOMP\end{tabular} \\
\cmidrule(r){1-1} \cmidrule(lr){2-2} \cmidrule(lr){3-3} \cmidrule(lr){4-4}
\texttt{VP}     & Order of object and verb                          & \begin{tabular}[c]{@{}l@{}}VT $\rightarrow$ (S$\mid$NP$_\text{SUBJ}$)\textbackslash{}NP$_\text{OBJ}$\\ VCOMP $\rightarrow$ (S$\mid$NP$_\text{SUBJ}$)\textbackslash{}SCOMP\\ REL $\rightarrow$ (NP$_\text{SUBJ}|$NP$_\text{SUBJ}$)$|$(S\textbackslash{}NP$_\text{OBJ}$)\end{tabular}                                              & \begin{tabular}[c]{@{}l@{}}VT $\rightarrow$ (S$\mid$NP$_\text{SUBJ}$)/NP$_\text{OBJ}$\\ VCOMP $\rightarrow$ (S$\mid$NP$_\text{SUBJ}$)/SCOMP\\ REL $\rightarrow$ (NP$_\text{SUBJ}$$|$NP$_\text{SUBJ}$)$|$(S/NP$_\text{OBJ}$)\end{tabular}                               \\ 
\cmidrule(r){1-1} \cmidrule(lr){2-2} \cmidrule(lr){3-3} \cmidrule(lr){4-4}
\texttt{O}     & Order of subject and object                        & Subject occurs before the object                                              & Object occurs before the subject                        \\ 
\cmidrule(r){1-1} \cmidrule(lr){2-2} \cmidrule(lr){3-3} \cmidrule(lr){4-4}
\texttt{COMP}   & Position of complementizer                     & COMP $\rightarrow$ SCOMP\textbackslash{}S                                                                                                                                                         & COMP $\rightarrow$ SCOMP/S                                                                                                                           \\ 
\cmidrule(r){1-1} \cmidrule(lr){2-2} \cmidrule(lr){3-3} \cmidrule(lr){4-4}
\texttt{PP}     & Postposition or preposition                         & PREP $\rightarrow$ (NP\textbackslash{}NP)/NP                                                                                                                                                      & PREP $\rightarrow$ (NP/NP)\textbackslash{}NP                                                                                                         \\ 
\cmidrule(r){1-1} \cmidrule(lr){2-2} \cmidrule(lr){3-3} \cmidrule(lr){4-4}
\texttt{ADJ}     & Order of adjective and noun & ADJ $\rightarrow$ NP/NP                                                                                                                                                                           & ADJ $\rightarrow$ NP\textbackslash{}NP                                                                                                               \\ 
\cmidrule(r){1-1} \cmidrule(lr){2-2} \cmidrule(lr){3-3} \cmidrule(lr){4-4}
\texttt{REL}    & Position of relativizer                            & REL $\rightarrow$ (NP$_\text{SUBJ}$/NP$_\text{SUBJ}$)\textbackslash{}(S$\mid$NP$_\text{OBJ}$)                                                                                                                   & REL $\rightarrow$ (NP$_\text{SUBJ}$\textbackslash{}NP$_\text{SUBJ}$)/(S$\mid$NP$_\text{OBJ}$)                                                                                     \\ 
\bottomrule
\end{tabular}
\caption{Binary word order parameters and their corresponding GCG categories. ``$\alpha\rightarrow\beta$'' indicates 
$\alpha$ is expanded to $\beta$
in the GCG derivation.
Some expansion rules interact with multiple word order parameters, e.g., VT $\rightarrow$ (S/NP$_\text{SUBJ}$)$\mid$NP$_\text{OBJ}$, and non-target directionalities are denoted as ``$\mid$'' representing either forward or backward slashes. 
}
\label{table:category_switch_derivations}
\end{table*}

\begin{table*}[t]

\centering
\small
\begin{tabularx}{\textwidth}{X X}
\toprule
GCG Lexical Syntactic Category & Example  \\
\cmidrule(lr){1-1} \cmidrule(lr){2-2}

\textbf{Noun Phrase (NP)} -- {\color{typegray}NP} 
& \colorbox{highlightbg}{\textbf{Kim}} ga kissed \colorbox{highlightbg}{\textbf{Sandy}} o \\

\textbf{Subject Marker} -- {\color{typegray}NP$_\text{SUBJ}$$\backslash$NP} 
& Kim \colorbox{highlightbg}{\textbf{ga}} kissed Sandy o \\

\textbf{Object Marker} -- {\color{typegray}NP$_\text{OBJ}$$\backslash$NP} 
& Kim ga kissed Sandy \colorbox{highlightbg}{\textbf{o}} \\

\textbf{Adjective (ADJ)} -- {\color{typegray}NP$|$NP} 
& \colorbox{highlightbg}{\textbf{red}} car ga ran \\

\textbf{Transitive Verb (VT)} -- {\color{typegray}(S$\mid$NP$_\text{SUBJ}$)$\mid$NP$_\text{OBJ}$} 
& Kim ga \colorbox{highlightbg}{\textbf{kissed}} Sandy o \\

\textbf{Intransitive Verb (VI)} -- {\color{typegray}S$\mid$NP$_\text{SUBJ}$} 
& red car ga \colorbox{highlightbg}{\textbf{ran}} \\

\textbf{Verb with Complement (VCOMP)} -- {\color{typegray}(S$\mid$NP$_\text{SUBJ}$)$\mid$SCOMP} 
& Kim ga \colorbox{highlightbg}{\textbf{believed}} that Sandy lied \\

\textbf{Complementizer (COMP)} -- {\color{typegray}SCOMP$\mid$S} 
& Kim ga believed \colorbox{highlightbg}{\textbf{that}} Sandy lied \\

\textbf{Preposition (PREP)} -- {\color{typegray}(NP$\mid$NP)$\mid$NP} 
& elf \colorbox{highlightbg}{\textbf{on}} shelf ga laughed \\

\textbf{Relativizer (REL)} -- {\color{typegray}(NP$_\text{SUBJ}$$\mid$NP$_\text{SUBJ}$)$\mid$(S$\mid$NP$_\text{OBJ}$)} 
& man ga \colorbox{highlightbg}{\textbf{whom}} I ga met laughed \\
\textbf{Conjunction} -- {\color{typegray}Var$\backslash$Var$/$Var} 
& Kim \colorbox{highlightbg}{\textbf{and}} Sandy ga ate \\
\bottomrule
\end{tabularx}
\caption{
Lexical syntactic categories, their derivations, and their examples (colored) supplemented with an English sentence as a context.
The vertical bars ``$\mid$'' in the GCG lexical syntactic categories represent either forward or backward slashes, determined by word order parameters listed in Table~\ref{table:category_switch_derivations}.
}
\label{table:category_descriptions}
\end{table*}

\section{Dataset}

\subsection{Overview}
We introduce a new set of ALs designed using our GCG framework, which allows us to create a wider range of ALs that reflect different word orders and long-distance dependencies. 
We generally replicate the experiments of 
\citet{DBLP:conf/acl/KuribayashiUYOB24} and 
\citet{DBLP:conf/acl/WhiteC20}
on our new datasets and extend these with generalization tests.
Because GCGs can, in principle, generate all syntactic patterns observed in natural languages, our framework offers a more comprehensive framework
to evaluate neural LMs. 
We illustrate this flexibility by extending the dataset of \citet{DBLP:conf/acl/WhiteC20} to include object relative clauses with potentially unbounded dependencies.
Our ALs are parameterized by word order parameters (Table~\ref{table:category_switch_derivations}), similarly to \citet{DBLP:conf/acl/WhiteC20}.
Each binary word order parameter controls the order of components within sets of constructions; for example, parameter \textsc{S} changes the order of subject and verb.
By setting these parameters independently and exhaustively, we create a set of ALs that differ in word order rules from each other.
All parameters except \texttt{O} follow those used by \citet{DBLP:conf/acl/WhiteC20}.
The additional \texttt{O} parameter introduces a subject–object ordering rule, which enables us to cover VSO and OSV word orders that were not represented in the ALs created by \citet{DBLP:conf/acl/WhiteC20}, resulting in 96 distinct ALs.

\subsection{Lexicon}
We first define 11 GCG lexical syntactic categories as shown in Table \ref{table:category_descriptions}.
 The directionality of the slashes in each category will be determined once the word order parameters are set (Table \ref{table:category_switch_derivations}).
Note that we include subject and object markers, and these consistently adopt postpositional case marking, following \citet{DBLP:conf/acl/WhiteC20}.\footnote{We changed the case marking system to be word-level (\textit{Taro -ga [whom I met]}) rather than phrase-level (\textit{Taro [whom I met] -ga}) adopted in~\citet{DBLP:conf/acl/WhiteC20}. Here, \textit{ga} is a nominal case marker.}
Our lexicon is the same size as that in \citet{DBLP:conf/acl/WhiteC20}, consisting mostly of English words.
To simplify the setting, we currently disregard subject-verb number agreement and trained word-level LMs without subword tokenization; that is, phonological and morphological patterns are not modeled by our LMs.

\subsection{Generating the Datasets}
To test the length generalization of LMs, we create three variations of the AL corpus: (i) \textsc{Short} with a length of 3--8 words, (ii) \textsc{Medium} with a length of 9--10 words, and (iii) \textsc{Long} with a length of 11--20 words.
Only the \textsc{Short} part is used for LM training, and the models are tested in held-out \textsc{Short}, \textsc{Medium}, and \textsc{Long} test sets.

The datasets are generated over several steps:
\begin{enumerate}
   
    \item \textbf{Set word order parameters.} 
    We generate the AL corpus for each combination of the seven parameters.
    Each AL is defined by a unique combination of parameter values, such as 0101101 for ``English''
    which corresponds to settings S=0, VP=1, O=0, COMP=1, PP=1, ADJ=0, and REL=1 
    (see Table~\ref{table:category_switch_derivations}).
    
    \item \textbf{Generate templates.} Once word order parameters are fixed, to ensure coverage of all valid sentences in each AL, we generate all possible sequences of word categories up to a length of 10 (for \textsc{Short} and \textsc{Medium} sets). 
    These category sequences are then parsed using a GCG parser configured according to the respective grammar with the lexical syntactic categories as terminal symbols.\footnote{We modify the NLTK CCGChartParser \cite{DBLP:books/daglib/0022921} by disabling type raising and incorporating the permutation operation described in \citet{briscoe1997co,briscoe2000grammatical}, which we use to parse our sentence templates.}
    A sequence of word categories is treated as grammatical if the parser produces at least one derivation with \texttt{S} as a root. 

    \item \textbf{Sample lexicons.} 
    Once grammatical templates with word categories are generated, we build sentences by randomly sampling lexicons for each word category.
    The number of sampled sentences is adjusted based on some policies.
    In our case, we sampled sentences to form a uniform distribution of sentence length within the training and test data, e.g., 1K of length-3 sentences, 1K of length-4 sentences, ..., 1K of length-8 sentences.
     An example of a valid sentence parse is illustrated in Figure~\ref{fig:derivation_sampling}.
   
    \item \textbf{Augument \textsc{Long} set.} Using the existing templates of lengths 3-10 words (\textsc{Short} and \textsc{Medium}), we create the templates for the \textsc{Long} test set, where the template lengths are 11-20 words. 
    We extend the existing templates in 3 different ways: 
    \begin{enumerate}
        \item \textbf{Concatenation:} 2 valid templates are concatenated end to end.
        \item \textbf{Mid-sentence insertion with a conjunction:} A conjunction and the second template are inserted at different points in the first template.
        \item \textbf{Appending with a conjunction:} Appending a template to another template using a conjunction.
    \end{enumerate}
    We filter the valid extended templates by parsing them using the GCG parsers for all 96 ALs, as previously done for the templates of length 3-10.
\item \textbf{Sample lexicons for templates of \textsc{Long} set.} 
We randomly sample 20,000 unique valid templates for each of the 96 ALs and, for each template, we sample one sentence from the lexicon. 
For each AL, we end up with 20,000 unique sentences of length 11-20. 

\end{enumerate}

Appendix~\ref{app:data} shows further details on the GCG parser configuration, as well as the statistics of the data we generated, including the template numbers for each length.

\begin{figure*}[t]
    \centering
    \begin{subfigure}{1.0\linewidth}
        \centering
        \includegraphics[width=1.0\linewidth]{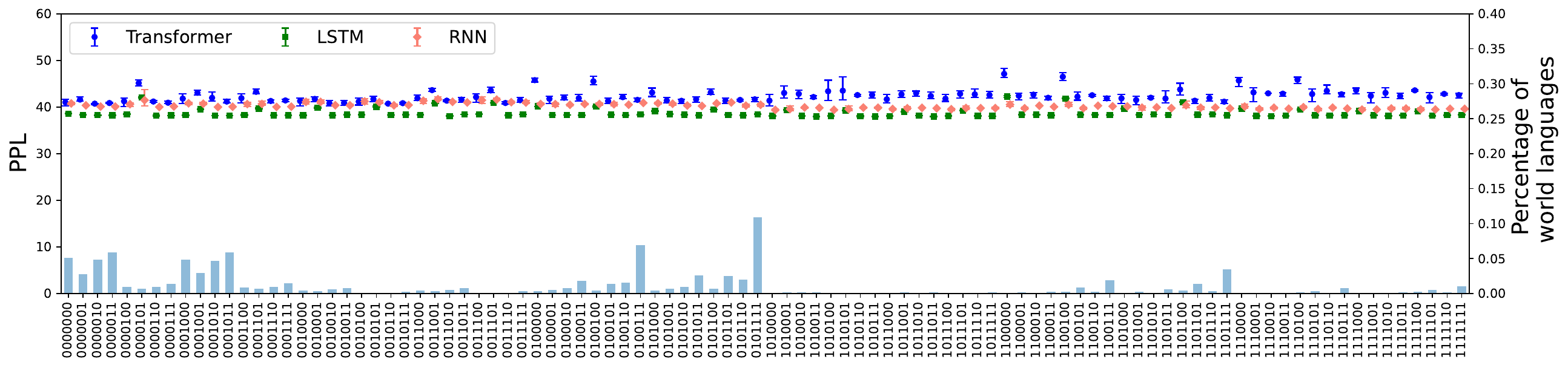}
        \caption{\textsc{Short} test (length 3--8).}
        \label{fig:perplexity-in-domain}
    \end{subfigure}

    \vspace{0.5em} 

    \begin{subfigure}{1.0\linewidth}
        \centering
        \includegraphics[width=1.0\linewidth]{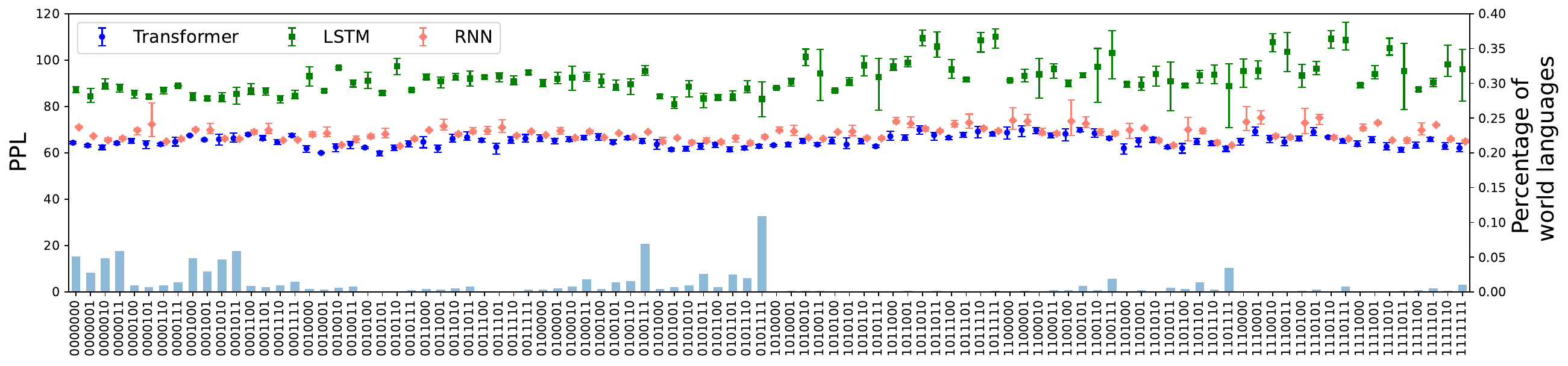}
        \caption{\textsc{Medium} test (length 9--10).}
        \label{fig:perplexity-out-domain}
    \end{subfigure}

    \begin{subfigure}{1.0\linewidth}
        \centering
        \includegraphics[width=1.0\linewidth]{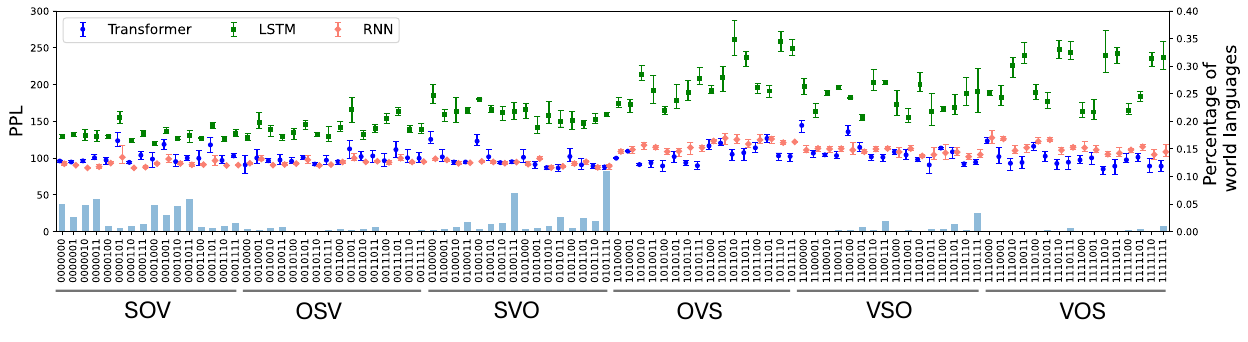}
        \caption{\textsc{Long} test (length 11--20).}
        \label{fig:perplexity-out-domain-long}
    \end{subfigure}
    
    \caption{Distributions of perplexities and typological plausibility across languages. The error bars indicate max and min PPLs within three runs.}
    \label{fig:perplexity-comparison}
\end{figure*}

\begin{table*}[t]
    \centering
    \scriptsize
    \tabcolsep 0.05cm
    \begin{tabular}{lrrrrrrrrrrrrrrrrrrrrr}
    \toprule
        & \multicolumn{7}{c}{\textsc{Short}} & \multicolumn{7}{c}{\textsc{Medium}} & \multicolumn{7}{c}{\textsc{Long}} \\
         \cmidrule(lr){2-8} \cmidrule(lr){9-15} \cmidrule(lr){16-22} 
        Model & SOV & OSV & SVO & OVS & VSO & VOS  & TA $\downarrow$ & SOV & OSV & SVO & OVS & VSO & VOS & TA $\downarrow$ & SOV & OSV & SVO & OVS & VSO & VOS & TA $\downarrow$ \\
        \cmidrule(lr){1-1} \cmidrule(lr){2-7} \cmidrule(lr){8-8} \cmidrule(lr){9-14} \cmidrule(lr){15-15} \cmidrule(lr){16-21} \cmidrule(lr){22-22}
        Transformer (PPL $\downarrow$) & 41.8 & 41.6 & 42.3 & 42.6 & 42.7 & 43.3
 & \textbf{$-$27.7}$^\dagger$  & 65.2 & 63.5 & 64.2 & 65.9 & 66.1 & 65.0 &
 \textbf{$-$10.4}$\:\:$ & 102.3 & 99.4 & 97.9 & 104.0 & 107.6 & 97.9 &  \textbf{$-$19.2}$\:\:$ \\
        LSTM (PPL $\downarrow$) & 38.7 & 38.8 & 38.7 & 38.4 & 39.1 & 38.5 & \textbf{$-$14.2}$\:\:$ & 85.9 & 91.7 & 88.0 & 97.5 & 92.9 & 97.9 & \textbf{$-$31.0}$^\dagger$ & 131.9 & 141.5 & 160.7 & 205.5 & 180.9 & 207.5 & \textbf{$-$33.4}$^\dagger$ \\
        RNN (PPL $\downarrow$) & 40.4 & 41.0 & 40.6 & 39.7 & 40.1 & 39.7 & 13.0$\:\:$ & 67.8 & 67.9 & 66.7 & 69.6 & 69.0 & 69.4 & \textbf{$-$17.4}$\:\:$ & 91.8 & 94.6 & 93.2 & 118.0 & 109.0 & 114.2 & \textbf{$-$43.1}$^\dagger$\\
        \cmidrule(lr){1-22}
        Natural Lang. (Prob. $\uparrow$) & 0.54 & 0.04 & 0.23 & 0.01 & 0.12 & 0.05 & - & 0.54 & 0.04 & 0.23 & 0.01 & 0.12 & 0.05 & - & 0.54 & 0.04 & 0.23 & 0.01 & 0.12 & 0.05 & - \\
        \bottomrule
    \end{tabular}
    \caption{Average PPLs within each base word order group as well as Pearson's correlation coefficient between PPL and the frequency of respective word order in the world. Negative TA (typological alignment) scores are highlighted in bold. Statistical significance of correlation coefficient (p<0.05) is marked with $\dagger$.}
    \label{tab:generalization}
\end{table*}

\begin{table*}
    \centering
    \small
    \begin{tabular}{lp{6cm}p{6cm}}
    \toprule
    Language & Recursive Relative Clauses & Embedded Relative Clause \\
    \cmidrule(lr){1-1} \cmidrule(lr){2-2} \cmidrule(lr){3-3} 
      0000000   & John ga promised which pasta ga nibbles which fruits ga wall o received & John ga pasta ga nibbles that said which fruits ga wall o received \\
      \cmidrule(lr){2-2} \cmidrule(lr){3-3} 
      0101101 (English) & fruits ga which pasta ga which John ga promised nibbles received wall o & fruits ga which John ga said that pasta ga nibbles received wall o \\
      \cmidrule(lr){2-2} \cmidrule(lr){3-3} 
      1111111   & received wall o fruits ga which nibbles pasta ga which promised John ga & received wall o fruits ga which said that nibbles pasta ga John ga \\
      \bottomrule
    \end{tabular}
    \caption{Examples in challenging test sets. The examples with the 0101101 word order parameters follow the basic English word order.}
    \label{tab:target_data}
\end{table*}

\begin{table}[t]
    \centering
    \small
    \tabcolsep 0.1cm
    \begin{tabular}{lrr}
    \toprule
        Model & \textsc{Recursive} (TA $\downarrow$) & \textsc{Embedded} (TA $\downarrow$) \\
        \cmidrule(lr){1-1} \cmidrule(lr){2-2} \cmidrule(lr){3-3} 
        Transformer & $-$5.1 & $-$23.5$^\dagger$ \\
        LSTM & 9.2 & $-$3.7$\:\:$ \\
        RNN & 12.9 & $-$18.1$^\dagger$ \\
        \bottomrule
    \end{tabular}
    \caption{Correlation between PPL in the targeted evaluation set for each language and typological plausibility. Statistical significance of correlation coefficient (p<0.05) is marked with $\dagger$.}
    \label{tab:targeted_generalization}
\end{table}
\section{Experimental Settings}

\paragraph{Models.}
We evaluate three variants of neural LMs: simple RNN~\cite{elman1990finding}, LSTM \cite{hochreiter1997long}, and Transformer \cite{vaswani2017attention}.
These models are trained using the Fairseq toolkit~\cite{ott2019fairseq}.
We quantify their inductive bias on what kind of word order they are good at for productive generalization.
See Appendix~\ref{app:model} for more details on the models.

\paragraph{Training.}
The training set consists of 80K sentences of lengths 3-8 words (\textsc{Short} training set).
The sentence length is equally distributed, and in each length, templates are also uniformly sampled.\footnote{Longer sentences have a larger number of grammatically possible templates; thus, this uniform sampling automatically introduces the tendency that shorter templates are more frequently selected.}
We stop the training based on an early-stopping criterion with a patience of five epochs (i.e., the training stops when validation loss does not decrease in five consecutive epochs) and a maximum of 10,000 update steps.

\paragraph{Evaluation.}
We use different evaluation metrics in different experiments, but they are all focused on generalization for longer sentences
($\geq 9$; \textsc{Medium} and \textsc{Long} sets)
than those in the training data.
We trained three LMs with different seeds for model initialization, and reported scores are the average of three runs.

\paragraph{Typological alignment (TA).}
We measure perplexity (PPL), the geometric mean of word probabilities across sequences, in each language.
That is, we obtained the PPL distribution over the 96 languages we used.
Following~\citet{DBLP:conf/acl/KuribayashiUYOB24}, we report Pearson's correlation coefficients between PPL and the percentage of respective word order in the world. 
The typological distribution is based on the percentage of languages that adopt the respective word order estimated with WALS~\cite{wals} and Grambank~\cite{grambank_dataset_zenodo_v1}.\footnote{We basically used the WALS database~\cite{wals} to count the frequency of word orders, following the same procedure as~\cite{DBLP:conf/acl/KuribayashiUYOB24}, and the COMP statistics are supplemented with Grambank~\cite{grambank_dataset_zenodo_v1} as the COMP statistics are not recorded in WALS.}
Lower negative correlation indicates that learnability is better aligned with typological commonality, i.e., the more common the word order is, the more easily the model learns and productively generalizes it.
Appendix~\ref{app:dartifacts} includes details on these databases.

\section{Experiment 1: PPLs}
\label{sec:exp1}

\paragraph{Evaluation settings.}
We first measure the PPLs on \textsc{Medium} and \textsc{Long} (out-of-domain) test sets with 20K longer sentences for each set.
For a comparison, we also measure PPLs on the 20K \textsc{Short} (in-domain) set with the same length distribution, i.e., length of 3--8, as training data.
We ensured that there is no overlap between each test set and training set, and all the vocabulary in the test sets is used at least once in the training set.
\vspace{-1.6mm}
\newline
\paragraph{Results.}
Figure~\ref{fig:perplexity-comparison} illustrates the PPL and typological distributions, and Table~\ref{tab:generalization} summarizes the results.
First, \textbf{the PPLs in longer test sets have larger variance}  (Figure~\ref{fig:perplexity-comparison}), and thus are more informative about which word order is easier to learn and generalize for a particular LM.
The flat PPL distribution in \textsc{Short} set, particularly for LSTMs, was also reported in~\citet{DBLP:conf/acl/WhiteC20}.
Second, Table~\ref{tab:generalization} shows that \textbf{the TA score and PPLs for each word order group substantially change between in-domain (\textsc{Short}) and out-of-domain (\textsc{Medium} and \textsc{Long}) evaluation.}
In particular, the TA scores in the out-of-domain evaluation are consistently negative, while not in the in-domain evaluation.\footnote{This negative result in in-domain data seems to contradict~\citet{DBLP:conf/acl/KuribayashiUYOB24}, where length generalization is not considered, but the advantage of working memory limitation is reported. This might be due to the difference in the length distribution of the training data (i.e., our study uses much shorter sentences than them), and we clarify that their results do not always hold if the training/evaluation domain is limited.}
These suggest that typologically plausible word orders tend to be easier for LMs to productively generalize, in contrast to just fitting to the in-domain data.
Third, we also have some architecture-dependent differences between in-domain and out-of-domain PPLs; specifically, \textbf{the TA scores for LSTM and RNN drastically improved in the out-of-domain evaluation, and RNN with \textsc{Long} test set achieved the best (lowest) correlation in all settings}.
In contrast, Transformer yielded a good correlation only with the in-domain test, which diminished in out-of-domain tests.
The out-of-domain results are somewhat intuitive if one believes that typologically common patterns are a consequence of human limited working memory, and the LSTM and RNN with recurrent model architecture have such cognitively plausible constraints, at least compared to the Transformer architecture.

\section{Experiment 2: PPLs in Targeted Generalization Sets}
\label{sec:exp2}
PPLs reported in~\cref{sec:exp1} are a holistic measure of out-of-domain generalization, given that the data is less focused on specific linguistic phenomena.

\paragraph{Evaluation settings.}
As a complementary evaluation, we introduce additional challenging out-of-domain test sets that focus on unbounded dependency constructions:  (i) recursive relative clauses, where two relative clauses are used in a nested; and (ii) embedded relative clauses, where the relative clause is in another subordinate clause, such as ``he said'' (Table~\ref{tab:target_data}).
We refer to these test sets as the \textsc{Recursive} and \textsc{Embedded} test sets, respectively.
All the sentences have the same construction as shown in Table~\ref{tab:target_data}, and lexicons are randomly sampled, resulting in 500 test sentences.
Note that these constructions are successfully regarded as grammatical under our GCG-based framework with the permutation operation, and are not included in the training set as they exceed the length of 8.
We report the TA score, i.e., correlation between PPL and typological distribution, on these challenging test sets.

\paragraph{Results.}
Table~\ref{tab:targeted_generalization} shows the TA scores for each challenging set and model.
In the \textsc{Recursive} set, the correlations are not statistically significant.
 The ease of such generalization was not related to the typological plausibility of word order, and possibly LMs simply failed to learn such a complex structure.
In the \textsc{Embedded} set, the correlations tend to be negative, and Transformer and RNN exhibited statistically significant correlations.
This result in \textsc{Embedded} set is overall consistent with the previous finding that typologically common ALs are easier to generalize for LMs.
That is, we found that, when the evaluation is extended to specific complex constructions, the results are somewhat phenomenon-dependent and require further investigation with broader-coverage targeted evaluations.

\begin{table*}
    \centering
    \small
    \begin{tabular}{lp{6.2cm}p{6.2cm}}
    \toprule
    Language & Case Type Judgment & Verb Type Judgment \\
    \cmidrule(lr){1-1} \cmidrule(lr){2-2} \cmidrule(lr){3-3} 
      \multirow{2}{*}{0000000}   & fluffy soft and intelligent mango ga owl o controls & green machine ga escorts which scooter ga walk \\
      & *fluffy soft and intelligent mango \underline{o} owl o controls & *green machine ga \underline{evolves} which scooter ga walk \\
      \cmidrule(lr){2-2} \cmidrule(lr){3-3} 
      \multirow{2}{*}{0101101 (English)} & fluffy soft and intelligent mango ga controls owl o & scooter ga which green machine ga escorts walk \\
      & *fluffy soft and intelligent mango \underline{o} controls owl o & *scooter ga which green machine ga \underline{evolves} walk \\
      \cmidrule(lr){2-2} \cmidrule(lr){3-3} 
      \multirow{2}{*}{1111111}   & controls owl o mango fluffy soft and intelligent ga & walk scooter ga which escorts machine green ga    \\
      & *controls owl o mango fluffy soft and intelligent \underline{o} & *walk scooter ga which \underline{evolves} machine green ga  \\
      \bottomrule
    \end{tabular}
    \caption{Examples in grammatical judgment tests. Ungrammatical sentences are marked with *. The examples with the 0101101 word order parameters follow the basic English word order.}
    \label{tab:grammaticality_data}
\end{table*}

\begin{table}[t]
    \centering
    \small
    \tabcolsep 0.1cm
    \begin{tabular}{lrrrr}
    \toprule
         & \multicolumn{2}{c}{Case Type} & \multicolumn{2}{c}{Verb Type}  \\
         \cmidrule(lr){2-3} \cmidrule(lr){4-5}
        Model & Corr.$\uparrow$ & Avg. Acc. & Corr.$\uparrow$ & Avg. Acc. \\
        \cmidrule(lr){1-1} \cmidrule(lr){2-2} \cmidrule(lr){3-3} \cmidrule(lr){4-4} \cmidrule(lr){5-5} 
        Transformer & 0.14$\:\:$ & 97.7$\pm$1.5 & 0.27$^\dagger$ & 81.0$\pm$14.7 \\
        LSTM & 0.03$\:\:$ & 97.2$\pm$1.4 & 0.28$^\dagger$ & 85.1$\pm$9.6 \\
        RNN & 0.21$^\dagger$ & 97.4$\pm$1.4 & 0.23$^\dagger$ & 77.4$\pm$15.5 \\
        \bottomrule
    \end{tabular}
    \caption{Correlation between accuracy and typological plausibility distribution, along with average and standard deviation of accuracy. Statistical significance of correlation coefficient (p<0.05) is marked with $\dagger$.}
    \label{tab:accuracy}
    \vspace{-2mm}
\end{table}

\section{Experiment 3: Grammaticality Judgment Accuracy}
\label{sec:exp3}
Lastly, we also perform grammaticality judgment evaluation as orthogonal to PPL evaluation, following the widely adopted minimal pair grammaticality judgment paradigm~\cite{warstadt-etal-2020-blimp-benchmark}.

\paragraph{Evaluation settings.}
As a case study, we selected two simple test cases: (i) case type accuracy, and (ii) verb type selection accuracy (Table~\ref{tab:grammaticality_data}).
Accuracy is measured based on whether a model could assign a high sentence probability to a grammatical sentence, given a pair of grammatical and ungrammatical ones. 
500 sentences are first sampled from the \textsc{Medium} set used in~\cref{sec:exp1} as out-of-domain grammatical sentences.
For each grammatical sentence, in the case type accuracy data, an ungrammatical option is created by replacing a case marker with a grammatically incorrect one (i.e., ga$\rightarrow$o or o$\rightarrow$ga).
In the verb type data, a transitive verb in the original sentence is wrongly replaced with an intransitive verb as an ungrammatical option  (e.g., escorts$\rightarrow$evolves).
The target token to be replaced is randomly selected if there are several candidates in a sentence. 
Note that, in our tests, the sentence length is aligned between the two options, and thus we simply calculated and compared the accumulated sentence probability $p(s)=\prod_{w_i\in s}p(w_i|\bm w_{<i})$ without any length normalization.
The grammatical judgment accuracy is measured in each word order configuration, and the correlation between the accuracies and typological plausibilities over 96 languages is reported (noted as Corr.).
This correlation should be positive if typologically common ALs are easier to learn.

\paragraph{Results.}
Table~\ref{tab:accuracy} shows the correlation scores, as well as average accuracy for each setting.
\textbf{All the correlations are positive, but only the RNN showed a statistically significant correlation in both settings.} 
These results are in line with the findings in~\cref{sec:exp1} that typologically frequent word order facilitates grammar acquisition, and a model with limited working memory yields better typological alignment.
To sum up all the experiments, the RNN exhibited superior typological alignment in length generalization, at least compared to the Transformer, especially in~\cref{sec:exp1} and~\cref{sec:exp3} (and somewhat comparable results in~\cref{sec:exp2}).
Given that RNN has the most limited working memory, as it does not have the gate mechanism of the LSTM or attention-based context access of a Transformer, this suggests that working memory limits create inductive bias predicting typological word order distributions.

\section{Conclusion}
In this paper, we create an AL framework inspired by \citet{DBLP:conf/acl/WhiteC20} to assess LM inductive biases towards different word orders.
We extend their framework from a PCFG to a GCG, and use 96 ALs to evaluate simple RNN, LSTM and Transformer LMs.
We calculate perplexity (PPL) on short, medium and long sentences, and observe a moderate alignment between PPL and frequency of word order in attested NLs, particularly in the out-of-domain evaluation with more complex linguistic constructions.
Overall, we observe that the performance of recurrent models, especially RNNs, provides good correlation with typological distributions, indicating that they may be the most typologically aligned models that generalize
effectively on typologically frequent word order patterns. 
In contrast, Transformers seem to be the least aligned when evaluated 
on generalization to longer sentences.

\section*{Limitations}
While our artificial language (AL) framework provides a controlled environment for evaluating language models (LMs), it does not fully capture the richness and variability of natural languages. 
The ALs used in this study are simplified and do not, for example, differentiate between verb tenses or include subject-verb agreement. 
We also do not explore ambiguity, and ensure that each word in the lexicon belongs to exactly one category, unlike in NLs.
\hidden{}
Future work is needed to systematically investigate a broader range of linguistic phenomena within this framework.

In the future, there are different avenues that we aim to explore. 
We would like to explore how different training methods can affect model learning and generalization.
Another potential future direction to explore is to investigate model learning and behavior when we introduce more features found in NLs, for example, subject-verb number agreement, or lexical ambiguity.

\section*{Ethical Statement}
The data used in this paper is artificially generated data that is based mostly on English. 
There is no sensitive information in the data, and no security risks in the contents of this paper.
We have no ethical concerns with the contents of this paper.

\section*{AI Writing/Coding Assistance Policy}
We occasionally used writing assistance systems, i.e., Grammarly and ChatGPT, but these are for the purpose of correcting grammatical/spelling errors and adjusting wording.
In other words, our use of AI writing assistance falls under the category (a) Assistance purely with the language of the paper, described in ARR.
\bibliography{acl_latex}

\appendix

\clearpage
\section{Dataset Details}
\label{app:data}
\subsection{Heuristics Applied During \textsc{Short} and \textsc{Medium} Template Generation}
To improve the efficiency of the template generation process, we apply a set of heuristics to filter out templates that would not produce valid sentences in any of our artificial languages. 

We discard templates that meet any of the following criteria:
\begin{enumerate}
    \item Contain fewer than 3 words (since all grammars require at least 3 words for a valid sentence),
    \begin{figure}
    \includegraphics[width=0.48\textwidth]{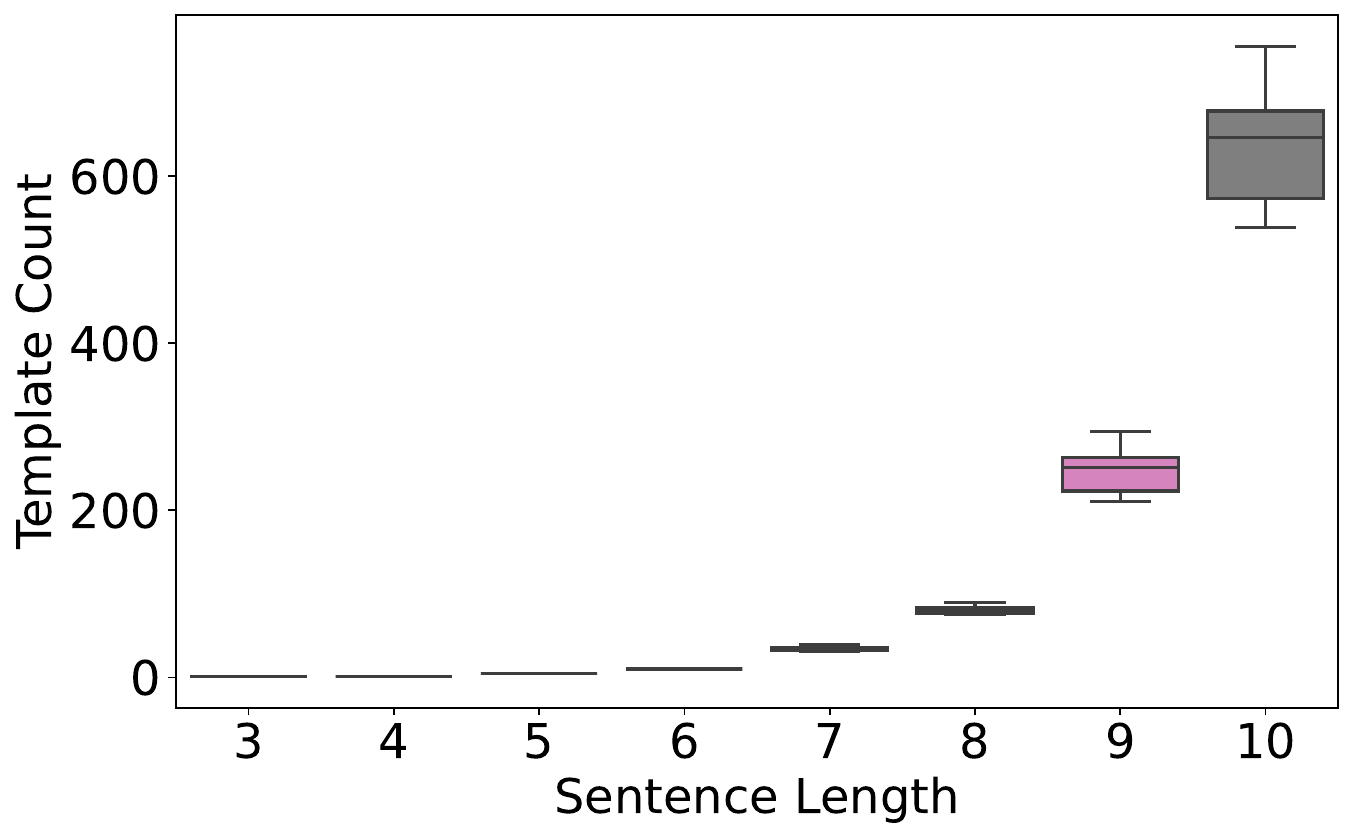}
    \caption{The distribution of the \textsc{Short} and \textsc{Medium} template lengths in our ALs (X-axis: template length, Y-axis: template count). The box and error bars present Q1 and Q3 percentiles, respectively.}
    \label{fig:distribution_template_lengths}
\end{figure}
\begin{figure}[t]
    \includegraphics[width=0.48\textwidth]{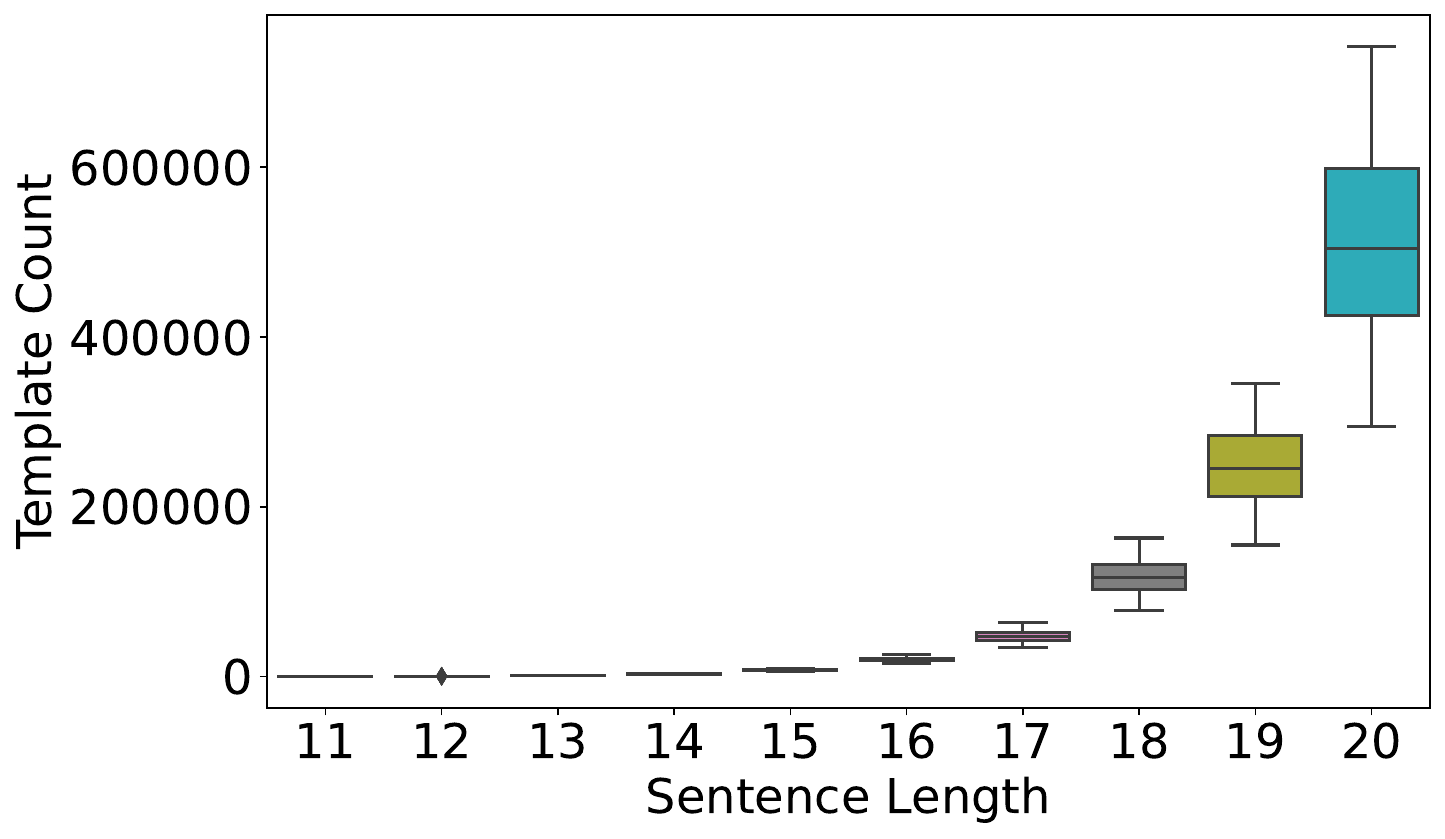}
    \caption{The distribution of all the \textsc{Long} template lengths in the extended templates created from the \textsc{Short} and \textsc{Medium} templates (X-axis: template length, Y-axis: template count). The box and error bars present Q1 and Q3 percentiles, respectively.}
    \label{fig:distribution_long_template_lengths}
\end{figure}

\begin{figure}
    \includegraphics[width=0.48\textwidth]{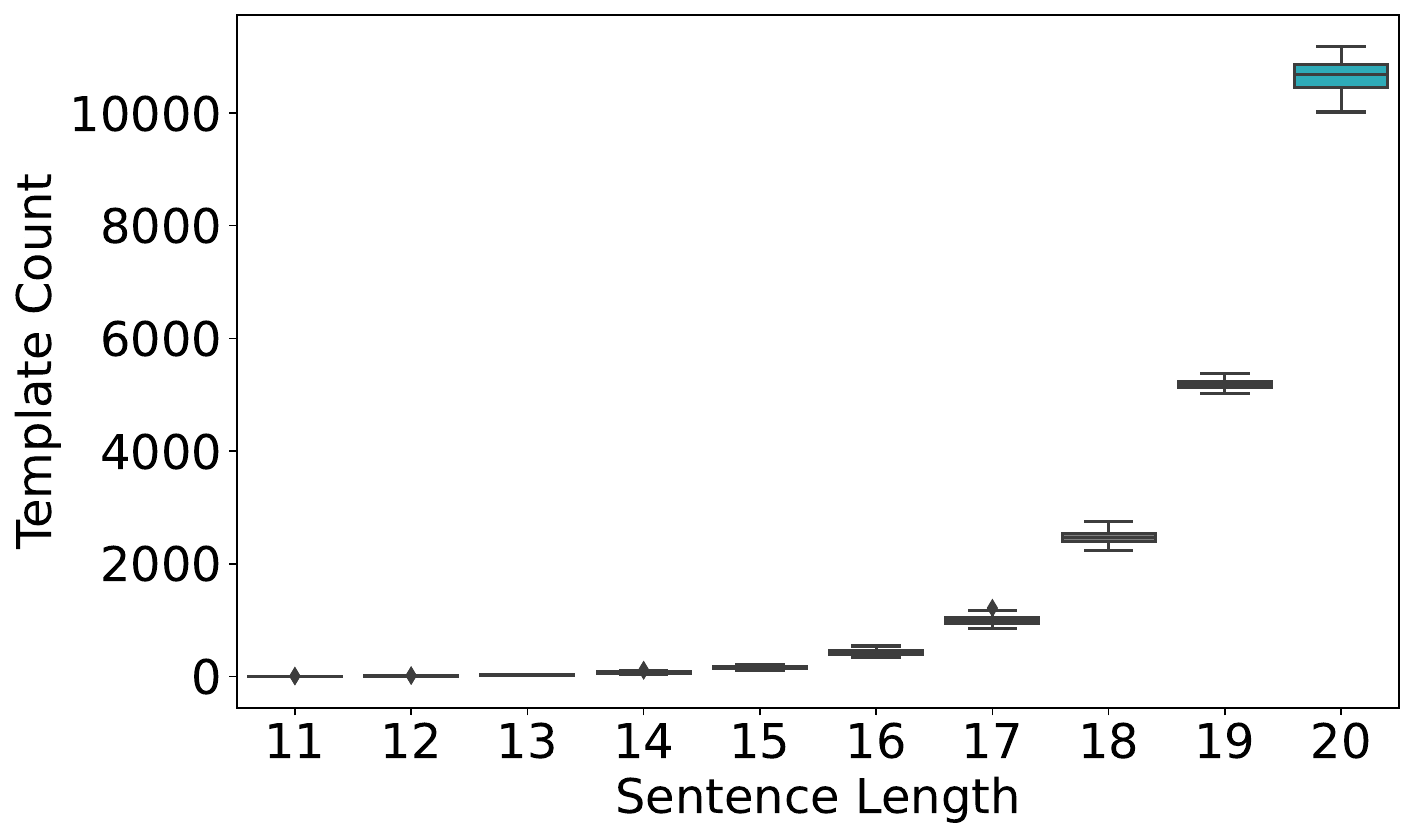}
    \caption{The distribution of the sampled \textsc{Long} template lengths, which we use to sample the sentences for the \textsc{Long} test set (X-axis: template length, Y-axis: template count). The box and error bars present Q1 and Q3 percentiles, respectively.}
    \label{fig:distribution_long_sampled_template_lengths}
\end{figure}
    \item Begin with a conjunction,
    \item End with a conjunction,
    \item Include two consecutive conjunctions,
    \item Include two consecutive prepositions,
    \item Start with subject or object markers,
    \item Contain more subject and object markers than NPs,
    \item Contain a complementiser without an associated complement verb.
\end{enumerate}

We plot the distribution of the different template lengths across our ALs.
We show this in Figure~\ref{fig:distribution_template_lengths} for \textsc{Short} and \textsc{Medium} length templates and Figures \ref{fig:distribution_long_template_lengths} and \ref{fig:distribution_long_sampled_template_lengths} for \textsc{Long} templates.
There is a slight variation in the number of templates in each AL, which is attributed to constraints naturally imposed by GCG, e.g., SVO word order can create ``S1 V1 and S2 V2 O'' as well as ``S V1 O1 and V2 O2'' structures but SOV word order can only create the ``S O1 V1 and O2 V2'' structure. 

\begin{table}[ht!]
\centering
\begin{tabular}{ll}
\toprule
\textbf{Category} & \textbf{Type} \\
\midrule
S &  Primitive\\
NP &   Primitive\\
VT & $(\text{S} \backslash \text{NPSUBJ})/\text{NPOBJ}$ \\
VI & $\text{S} \backslash \text{NPSUBJ}$ \\
VCOMP & $(\text{S} \backslash \text{NPSUBJ})/\text{SCOMP}$ \\
COMP & $\text{SCOMP}/\text{S}$ \\
PREP & $(\text{NP}/\text{NP})\backslash\text{NP}$ \\
ADJ & $\text{NP}/,\text{NP}$ \\
REL & $(\text{NPSUBJ} \backslash \text{NPSUBJ})/(\text{S}/\text{NPOBJ})$ \\
SUBJ & $\text{NPSUBJ} \backslash \text{NP}$ \\
OBJ & $\text{NPOBJ} \backslash \text{NP}$ \\
CONJ & $\text{var} \backslash ., @\text{var}/., @\text{var}$ \\
\bottomrule
\end{tabular}
\caption{GCG grammar for the word order consistent with the English language}
\label{tab:gcg_grammar_english}
\end{table}

\begin{table*}[ht!]
    \centering
    \small
    \begin{tabular}{lll}
    \toprule
        Artifact & License & Purpose \\
        \cmidrule(lr){1-1} \cmidrule(lr){2-2} \cmidrule(lr){3-3} 
        NLTK \cite{DBLP:books/daglib/0022921} & Apache License 2.0 & to parse sentences in data generation \\
        Fairseq~\cite{ott2019fairseq} & MIT Linense & to train LMs \\
        \citet{DBLP:conf/acl/WhiteC20} Data & MIT License & to determine dataset configuration\\
        WALS \cite{wals} & Creative Commons CC-BY 4.0 & to find word order statistics in NLs\\
        Grambank \cite{grambank_dataset_zenodo_v1} & Creative Commons CC-BY 4.0 & to find word order statistics in NLs\\
    \bottomrule
    \end{tabular}
    \caption{Details on artifacts we used in this study}
    \label{tab:artifacts}
\end{table*}

\begin{table*}[ht!]
    \centering
\fontsize{8}{9}\selectfont
\setlength{\tabcolsep}{2pt}
\begin{tabular}{ll}
\toprule
S & 82A Order of Subject and Verb~\cite{wals-82} \\
VP & 83A Order of Object and Verb~\cite{wals-83} \\
O & 81A Order of Subject, Object and Verb~\cite{wals-81} \\
\multirow{2}{*}{COMP} & Feature GB421: Is there a preposed complementizer in complements of verbs of thinking and/or knowing?~\cite{grambank_dataset_zenodo_v1} \\
& Feature GB422: Is there a postposed complementizer in complements of verbs of thinking and/or knowing?~\cite{grambank_dataset_zenodo_v1} \\
PP & 85A Order of Adposition and Noun Phrase~\cite{wals-85} \\
ADJ & 87A Order of Adjective and Noun~\cite{wals-87} \\
REL & 90A Order of Relative Clause and Noun~\cite{wals-90} \\
\bottomrule
\end{tabular}
\caption{WALS and Grambank chapters we used}
\label{tbl:wals}
\end{table*}

\subsection{Parser Configuration}
To parse templates and assign them to compatible artificial languages (ALs), we adapt the NLTK CCGChartParser \cite{DBLP:books/daglib/0022921}. 
We disable type raising, an operation available in Combinatory Categorial Grammar (CCG) \cite{steedman1996surface}, and instead implement the permutation rule described by \citet{briscoe1997co,briscoe2000grammatical}, which is part of Generalized Categorial Grammar (GCG) \cite{wood2014categorial}.

The NLTK CCGChartParser allows us to enforce parsing constraints: placing a comma, period, or underscore before a grammar argument disables composition, crossing, or substitution, respectively. 
We extend this by introducing a new symbol “@” to block permutation.

In our grammar definitions, we limit permutation to categories that function as verb functors (i.e., those involving S). 
We also constrain subject and object markers so that they only combine with NPs by disabling composition in the definitions of the NP$_\text{SUBJ}$ and NP$_\text{OBJ}$ categories.

GCG enables flexible word orders via permutation, which can cause overlap between word orders, for instance, OSV structures appearing in SOV datasets, or VSO in VOS datasets. 
To maintain clearer distinctions between word orders, we disable permutation for verbs when parsing templates for OSV, SOV, VOS, and OVS languages, except in cases where a relativizer category (REL) is present.

We provide an example of the SVO grammar that corresponds to English in Table~\ref{tab:gcg_grammar_english}.

\section{Information relevant to responsibility checklist}
\label{app:responsibility}

\subsection{Model Details}
\label{app:model}
We used exactly the same model hyperparameters as~\citet{DBLP:conf/acl/KuribayashiUYOB24} for Transformer, LSTM, and RNNs (see Table~\ref{tbl:hyper_params}).
These models are trained with the Fairseq toolkit~\cite{ott2019fairseq}.
We did not apply any subword tokenization, in contrast to~\citet{DBLP:conf/acl/KuribayashiUYOB24}, as we disregard morphological number agreement between subjects and verbs.
The whole model training and evaluation will be completed with approximately 150 GPU hours.

\subsection{Artifacts}
\label{app:dartifacts}
Table~\ref{tab:artifacts} shows all the artifacts we used, which follow the original intended use.
Specifically, NLTK is used for language analysis, fairseq is used for model training, and the linguistic database is used for accessing language statistics.
Table~\ref{tbl:wals} shows the exact chapters/features of linguistic databases we used.

\begin{table*}[ht]
    \centering
    \small
\begin{minipage}[t]{\hsize}
\renewcommand{\arraystretch}{0.4}
    \centering
    \begin{tabular}{p{3cm}p{5cm}p{4.5cm}} \toprule
     \multirow{10}{1cm}{Fairseq model}
      & share-decoder-input-output-embed & True \\
      & embed\_dim & 128 \\
      & ffn\_embed\_dim & 512 \\
      & layers & 2 \\
      & heads & 2 \\
      & dropout & 0.3 \\
      & attention\_dropout & 0.1 \\
      & \#params. & 462K \\
    \cmidrule(lr){1-1} \cmidrule(lr){2-2} \cmidrule(lr){3-3}
    \multirow{5}{*}{Optimizer} & algorithm & AdamW \\
    & learning rates & 5e-4 \\
    & betas & (0.9, 0.98) \\
    & weight decay & 0.01 \\
    & clip norm & 0.0 \\
    \cmidrule(lr){1-1} \cmidrule(lr){2-2} \cmidrule(lr){3-3}
    \multirow{3}{3cm}{Learning rate scheduler} & type & inverse\_sqrt \\
    & warmup updates & 400 \\
    & warmup init learning rate & 1e-7 \\
    \cmidrule(lr){1-1} \cmidrule(lr){2-2} \cmidrule(lr){3-3}
    \multirow{4}{*}{Training}   
    & batch size & 2,048 tokens \\
    & tokens-per-sample & 128 tokens \\
    & sample-break-mode & none \\
    & epochs & 10 \\ 
    \bottomrule
        \end{tabular}
        \subcaption{Transformer.}
        \label{tbl:hyperparam_tl}
        \vspace{0.2cm}
\end{minipage}

\begin{minipage}[t]{\hsize}
\renewcommand{\arraystretch}{0.4}
    \centering
    \begin{tabular}{p{3cm}p{5cm}p{4.5cm}} \toprule
     \multirow{8}{1cm}{Fairseq model}
      & share-decoder-input-output-embed & True \\
      & embed\_dim & 128 \\
      & hiden\_size & 512 \\
      & layers & 2 \\
      & dropout & 0.1 \\
      & \#params. & 3,547K \\
    \cmidrule(lr){1-1} \cmidrule(lr){2-2} \cmidrule(lr){3-3}
    \multirow{5}{*}{Optimizer} & algorithm & AdamW \\
    & learning rates & 5e-4 \\
    & betas & (0.9, 0.98) \\
    & weight decay & 0.01 \\
    & clip norm & 0.0 \\
    \cmidrule(lr){1-1} \cmidrule(lr){2-2} \cmidrule(lr){3-3}
    \multirow{3}{3cm}{Learning rate scheduler} & type & inverse\_sqrt \\
    & warmup updates & 400 \\
    & warmup init learning rate & 1e-7 \\
    \cmidrule(lr){1-1} \cmidrule(lr){2-2} \cmidrule(lr){3-3}
    \multirow{4}{3cm}{Training} & batch size & 2,048 tokens \\
    & tokens-per-sample & 128 tokens \\
    & sample-break-mode & none \\ 
    & epochs & 10 \\ \bottomrule
        \end{tabular}
        \subcaption{LSTM.}
        \label{tbl:hyperparam_lstm}
        \vspace{0.2cm}
\end{minipage}

\begin{minipage}[t]{\hsize}
\renewcommand{\arraystretch}{0.4}
    \centering
    \begin{tabular}{p{3cm}p{5cm}p{4.5cm}} \toprule
     \multirow{8}{1cm}{Fairseq model}
      & share-decoder-input-output-embed & True \\
      & embed\_dim & 64 \\
      & hiden\_size & 64 \\
      & layers & 2 \\
      & dropout & 0.1 \\
      & \#params. & 49K \\
    \cmidrule(lr){1-1} \cmidrule(lr){2-2} \cmidrule(lr){3-3}
    \multirow{5}{*}{Optimizer} & algorithm & AdamW \\
    & learning rates & 5e-4 \\
    & betas & (0.9, 0.98) \\
    & weight decay & 0.01 \\
    & clip norm & 0.0 \\
    \cmidrule(lr){1-1} \cmidrule(lr){2-2} \cmidrule(lr){3-3}
    \multirow{3}{3cm}{Learning rate scheduler} & type & inverse\_sqrt \\
    & warmup updates & 400 \\
    & warmup init learning rate & 1e-7 \\
    \cmidrule(lr){1-1} \cmidrule(lr){2-2} \cmidrule(lr){3-3}
    \multirow{4}{3cm}{Training} & batch size & 2,048 tokens \\
    & tokens-per-sample &  128 tokens \\
    & sample-break-mode & none \\ 
    & epochs & 10 \\ \bottomrule
        \end{tabular}
        \subcaption{RNN.}
        \label{tbl:hyperparam_rnn}
        \vspace{0.2cm}
\end{minipage}
\caption{Hyperparameters of LMs}
\label{tbl:hyper_params}
\end{table*}

\end{document}